\documentclass{IEEEtran}
\pdfoutput=1
\usepackage{tikz}
\usepackage{pgfplots}
\usepackage{verbatim}
\usepackage{aircraftshapes}
\pgfplotsset{compat=newest}
\pgfplotsset{every axis legend/.append style={%
		cells={anchor=west}}
}
\pgfplotsset{every y tick label/.append style={font=\footnotesize}}
\pgfplotsset{every x tick label/.append style={font=\footnotesize}}
\pgfplotsset{every axis x label/.append style={font=\footnotesize}}
\pgfplotsset{every axis y label/.append style={font=\footnotesize}}
\pgfplotsset{every axis legend/.append style={font=\footnotesize}}
\pgfplotsset{every axis title/.append style={font=\footnotesize}}
\usepgfplotslibrary{polar}
\usetikzlibrary{arrows}
\tikzset{>=stealth'}
\usepgfplotslibrary{groupplots}
\usepackage[per-mode=symbol,detect-all]{siunitx}
\usepackage{graphicx}
\DeclareGraphicsExtensions{.pdf, .jpg, .png}
\usepackage{amsmath}
\usepackage[noend]{algorithmic}
\usepackage{algorithm}
\usepackage{array}
\usepackage{fixltx2e}
\usepackage{float}
\usepackage{url}
\usepackage{booktabs}
\usepackage[capitalize]{cleveref}
\usepackage{todonotes}
\newcommand{\argmax}{\operatornamewithlimits{arg\,max}}

\makeatletter
\def\blx@maxline{77}
\makeatother

\title{Distributed Wildfire Surveillance with Autonomous Aircraft using Deep Reinforcement Learning}
\author{Kyle D. Julian$^{1}$ and Mykel J. Kochenderfer$^{2}$
	\thanks{$^{1}$Kyle D. Julian is a Ph.D. candidate in the Department of Aeronautics and Astronautics,
		Stanford University, Stanford, CA, 94305
		{\tt\small kjulian3@stanford.edu}}%
	\thanks{$^{2}$Mykel J. Kochenderfer is an Assistant Professor in the Department of Aeronautics and Astronautics,
		Stanford University, Stanford, CA, 94305
		{\tt\small mykel@stanford.edu}}%
}

\begin{document}
	\maketitle
	\thispagestyle{empty}
	\pagestyle{empty}
	
\begin{abstract}
	Teams of autonomous unmanned aircraft can be used to monitor wildfires, enabling firefighters to make informed decisions. However, controlling multiple autonomous fixed-wing aircraft to maximize forest fire coverage is a complex problem. The state space is high dimensional, the fire propagates stochastically, the sensor information is imperfect, and the aircraft must coordinate with each other to accomplish their mission. 
	This work presents two deep reinforcement learning approaches for training decentralized controllers that accommodate the high dimensionality and uncertainty inherent in the problem. The first approach controls the aircraft using immediate observations of the individual aircraft. The second approach allows aircraft to collaborate on a map of the wildfire's state and maintain a time history of locations visited, which are used as inputs to the controller. Simulation results show that both approaches allow the aircraft to accurately track wildfire expansions and outperform an online receding horizon controller. Additional simulations demonstrate that the approach scales with different numbers of aircraft and generalizes to different wildfire shapes.
\end{abstract}

\maketitle

\section{Introduction \label{sec:Intro}}

	 Wildfires consumed 10.1 million acres of land in 2015 and caused an estimated \$6 billion of damage from 1995 to 2014 \cite{FireCenter}. Having accurate information about the current state of a wildfire during the course of its evolution is important in deciding where to use fire suppressants and where to remove fuel. There are many approaches to monitoring wildfire growth including high-fidelity computer simulation \cite{finney1998farsite,rothermel1972mathematical,vacchiano2015implementation} and satellite images \cite{martin1999fire}, but these approaches do not offer real-time and high-resolution wildfire tracking. Aircraft have been used for decades to monitor wildfires due to their maneuverability and flexibility \cite{allison2016airborne}. Because deploying manned aircraft over fire can be dangerous and expensive, remotely-piloted unmanned aerial vehicles (UAVs) can have major safety and economic benefits for wildfire monitoring, and flight tests with the NASA Ikhana unmanned aircraft demonstrated the capabilities of UAVs in wildfire monitoring \cite{ambrosia2011ikhana}. To further reduce operational costs of wildfire surveillance, this work proposes using a team of small autonomous UAVs, which cost less to build and require no human control.
	 
	 To enable UAVs to autonomously monitor wildfires, sensor observations must be processed into meaningful information in real time. Merino et al. use feature matching techniques in flight tests with small autonomous helicopters with real controlled fires, and a central task planner was used to make decisions \cite{merino2012unmanned}. They illustrate feature matching techniques can be used to track the front of a fire from image data. Kukreti et al. demonstrate an algorithm for vision-based localization using a UAV that distinguishes fire pixels from non-fire pixels with great accuracy \cite{kukreti2016detection}. Yuan et al. applies image processing techniques such as filtering, color space conversion, and threshold segmentation to detect and track fires \cite{yuan2015uav}. De Vivo et al. use active contours for segmenting wildfire images with high accuracy \cite{de2018real}. These works show that raw images of fire can be converted into maps of wildfires to be used in UAV control.
	 
	 In addition, an accurate wildfire simulation model is required to develop and test wildfire monitoring UAV controllers. High-fidelty models incorporate variables such as fuel bulk density, fuel moisture, and specific heats to mathematically model the rate of spread for wildfires \cite{finney1998farsite,rothermel1972mathematical,vacchiano2015implementation}. Other approaches use stochastic wildfire models that account for wind and fuel variability \cite{Bertsimas2017,boychuk2009stochastic}. The work presented here uses a stochastic model inspired by these approaches.
	 
	 Given that wildfire locations can be extracted from images, one method for controlling UAVs tracks the outer edge of the fire using a waypoint planner. Casbeer et al. simulate a wildfire and track the fire boundary with multiple UAVs that turn around upon meeting another UAV \cite{casbeer2005forest}. This approach was extended by using cooperation constraint and rendezvous points to improve the distribution of aircraft around the wildfire perimeter \cite{beard2006decentralized}. Further work incorporates dynamic numbers of UAVs and monitoring moving boundaries \cite{cao2006dynamic,kingston2008decentralized}.
	 
	 Other methods take an information-theoretic approach. Paull et al. describe a method to survey an entire area using fixed-wing UAVs \cite{paull2014sensor}. A coverage map is used with a path planner to maximize the information gained during flight, and the authors prove that full coverage can be guaranteed over any desired area. Another approach leverages modern computer science algorithms to improve the decision making process when monitoring wildfires. Bertsimas et al. frame the problem of distributing aircraft for wildfire surveillance as a problem of dynamic resource allocation \cite{Bertsimas2017}. They show that Monte Carlo tree search and mathematical optimization surpass a baseline algorithm for monitoring wildfires.	 
	 
	 Controlling aircraft using information from images is a high-dimensional control problem. One technique for such problems is deep reinforcement learning (DRL). Google's DeepMind team demonstrated the effectiveness of DRL in creating intelligent agents to play Atari games and Go at super-human levels \cite{mnih2015human,silver2016mastering}. The strength of DRL lies in its ability to create intelligent agents without any expert domain knowledge. The agents are given the freedom to determine their own actions as functions of images, and over time the agent can learn to choose actions that outperform human experts. The work by DeepMind suggests that DRL could be effective for the wildfire problem. The algorithm learns a policy to maximize the accumulation of reward given images, leading to precise control in high dimensional state spaces.
	 
	 Previous approaches to autonomous wildfire surveillance separate the control system from the wildfire observations. However, wildfires can grow in unpredictable ways, making planning trajectories difficult. In addition, feature extraction limits the amount of information the controller can use to plan trajectories, and hand-tuned controllers using image features may not generalize well among all possible images. The contribution of this work is to use images gathered from wildfires to generate real-time bank angle commands that guide the aircraft around a wildfire as it expands. This method scales to multiple aircraft and allows them to efficiently work together to monitor the wildfire growth.
	 
	 This paper is organized as follows. A mathematical framework for both wildfire propagation and trajectory optimization is presented in \cref{sec:Formulation}. \Cref{sec:Approach1} details an approach to control aircraft based on immediate observations, while \cref{sec:Approach2} explains a second approach to collaboratively minimize the uncertainty of the wildfire's location. \Cref{sec:Solution} describes the solution method with deep reinforcement learning while \Cref{sec:Baseline} details a baseline receding horizon approach. \Cref{sec:Results} shows simulation results from both approaches and presents evaluation metrics that highlight their strengths and weaknesses.

\section{POMDP Approach \label{sec:Formulation}}
A Markov decision process is an optimization framework for sequential decision problems. Given an agent in state $s \in S$ taking action $a \in A$, the agent will transition to new state $s' \in S$ and receive reward $r \in R$. In this application, the agent is the aircraft being controlled, and the actions will dictate the aircraft's trajectory. Because the full state of the wildfire is unknown, decisions must be made using observations rather than the full state, making this problem a partially observable Markov decision process (POMDP). This section explains the POMDP formulation including the underlying wildfire mode, aircraft dynamics, and POMDP formulation



\subsection{Wildfire Model}
A stochastic wildfire model is needed to create a simulation environment to train and evaluate a controller. A stochastic fire propagation model similar to previous work was used \cite{Bertsimas2017}. First, a \SI{1}{\kilo\meter\squared} area of land is discretized into cells, forming a rectangular $100\times100$ grid. This area of land is small compared to real wildfires, but this approach can be easily scaled to larger areas. Each cell has two variables of interest:
\begin{enumerate}
	\item $F(s)$: Amount of burnable fuel remaining in cell $s$
	\item $B(s)$: Boolean variable signaling that cell $s$ is burning
\end{enumerate}

Each time the wildfire is updated, the cells can change in three ways. Burning cells will deplete more of their fuel as the fire consumes fuel. When the fuel reaches zero, the fire extinguishes within that cell. For non-burning cells with some fuel remaining, there is a probability $p(s)$ of igniting based on proximity to burning cells. Defining the burning rate as $\beta$, the wildfire propagation equations are
\begin{equation}
F_{t+1}(s) = 
\begin{cases}
\max(0,F_t(s)-\beta)& \text{if }  B_t(s)\\
F_t(s)              & \text{otherwise}
\end{cases}
\end{equation}	
	
\begin{equation} \label{eq:probIgnite}
p(s) = 
\begin{cases}
1 -\prod_{s'} (1-P(s,s')B_t(s')) & \text{if } F_t(s)>0 \\
0                         & \text{otherwise}
\end{cases}
\end{equation} 
where $P(s,s')$ represents the probability that cell $s'$ ignites cell $s$ in the next time step. $P(s,s') \propto d(s,s')^{-2}$ where $d(s,s')$ is the distance between the cells, so greater separation decreases the probability of fire spreading. Allowing a wildfire to spread beyond the immediate neighbors mimics the way embers can be cast farther away and ignite fires. To reduce computation time while allowing fire to spread beyond immediate neighbors, $P(s,s')=0$ for cells that are more than two cells away.

Simulation begins by randomly assigning fuel quantities to each cell and seeding the fire with some initial burning cells. Given that the amount of fuel in each cell has arbitrary units, let the burning rate $\beta$ equal one. With each consecutive step in the simulation, the wildfire grows as an expanding ring of fire while the burning cells use their fuel and eventually extinguish. One of the largest sources of variability in wildfire propagation is wind, which can be modeled by biasing the probability of cells igniting based upon the direction and strength of the wind. 

\begin{figure*}
	
	\begin{tikzpicture}
	\matrix [anchor=base] {
		\node {\quad \qquad \footnotesize No wind\qquad\qquad \qquad \qquad \qquad}; & \node {\quad \qquad \qquad \qquad \qquad \footnotesize Wind from west to east};\\
	};
	\end{tikzpicture}
	\centering
	\begin{tikzpicture}[]
\begin{groupplot}[width={5.0cm},height={5.0cm}, group style={horizontal sep=0.3cm, vertical sep=0.3cm, group size=4 by 4}]

\nextgroupplot [ylabel = {$T = \SI{0}{\second}$},title = {Fire Locations}, ticks=none,enlargelimits = false, axis on top]\addplot [point meta min=0, point meta max=22] graphics [xmin=10, xmax=1000, ymin=10, ymax=1000] {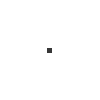};

\nextgroupplot [title = {Fuel Map}, ticks=none, enlargelimits = false, axis on top]\addplot [point meta min=0, point meta max=22] graphics [xmin=10, xmax=1000, ymin=10, ymax=1000] {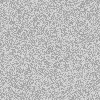};

\nextgroupplot [ title = {Fire Locations}, ticks=none,, enlargelimits = false, axis on top]\addplot [point meta min=0, point meta max=22] graphics [xmin=10, xmax=1000, ymin=10, ymax=1000] {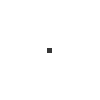};

\nextgroupplot [ title = {Fuel Map}, ticks=none, enlargelimits = false, axis on top]\addplot [point meta min=0, point meta max=22] graphics [xmin=10, xmax=1000, ymin=10, ymax=1000] {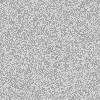};

\nextgroupplot [ylabel = {$T = \SI{50}{\second}$}, ticks=none,enlargelimits = false, axis on top]\addplot [point meta min=0, point meta max=22] graphics [xmin=10, xmax=1000, ymin=10, ymax=1000] {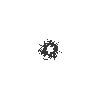};

\nextgroupplot [ ticks=none, enlargelimits = false, axis on top]\addplot [point meta min=0, point meta max=22] graphics [xmin=10, xmax=1000, ymin=10, ymax=1000] {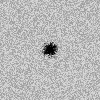};

\nextgroupplot [ticks=none,, enlargelimits = false, axis on top]\addplot [point meta min=0, point meta max=22] graphics [xmin=10, xmax=1000, ymin=10, ymax=1000] {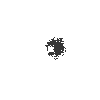};

\nextgroupplot [ticks=none, enlargelimits = false, axis on top]\addplot [point meta min=0, point meta max=22] graphics [xmin=10, xmax=1000, ymin=10, ymax=1000] {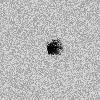};

\nextgroupplot [ylabel = {$T = \SI{100}{\second}$}, ticks=none,enlargelimits = false, axis on top]\addplot [point meta min=0, point meta max=22] graphics [xmin=10, xmax=1000, ymin=10, ymax=1000] {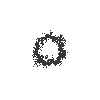};

\nextgroupplot [ticks=none, enlargelimits = false, axis on top]\addplot [point meta min=0, point meta max=22] graphics [xmin=10, xmax=1000, ymin=10, ymax=1000] {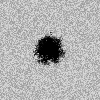};

\nextgroupplot [ticks=none,, enlargelimits = false, axis on top]\addplot [point meta min=0, point meta max=22] graphics [xmin=10, xmax=1000, ymin=10, ymax=1000] {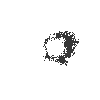};

\nextgroupplot [ ticks=none, enlargelimits = false, axis on top]\addplot [point meta min=0, point meta max=22] graphics [xmin=10, xmax=1000, ymin=10, ymax=1000] {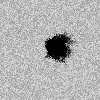};

\nextgroupplot [ylabel = {$T = \SI{150}{\second}$}, ticks=none,enlargelimits = false, axis on top]\addplot [point meta min=0, point meta max=22] graphics [xmin=10, xmax=1000, ymin=10, ymax=1000] {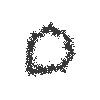};

\nextgroupplot [ticks=none, enlargelimits = false, axis on top]\addplot [point meta min=0, point meta max=22] graphics [xmin=10, xmax=1000, ymin=10, ymax=1000] {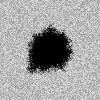};

\nextgroupplot [ticks=none,, enlargelimits = false, axis on top]\addplot [point meta min=0, point meta max=22] graphics [xmin=10, xmax=1000, ymin=10, ymax=1000] {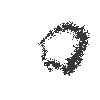};

\nextgroupplot [ ticks=none, enlargelimits = false, axis on top]\addplot [point meta min=0, point meta max=22] graphics [xmin=10, xmax=1000, ymin=10, ymax=1000] {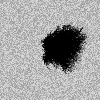};
\end{groupplot}

\end{tikzpicture}
	\caption{Wildfire propagation over time}
	\label{fig:FireProp}
\end{figure*}
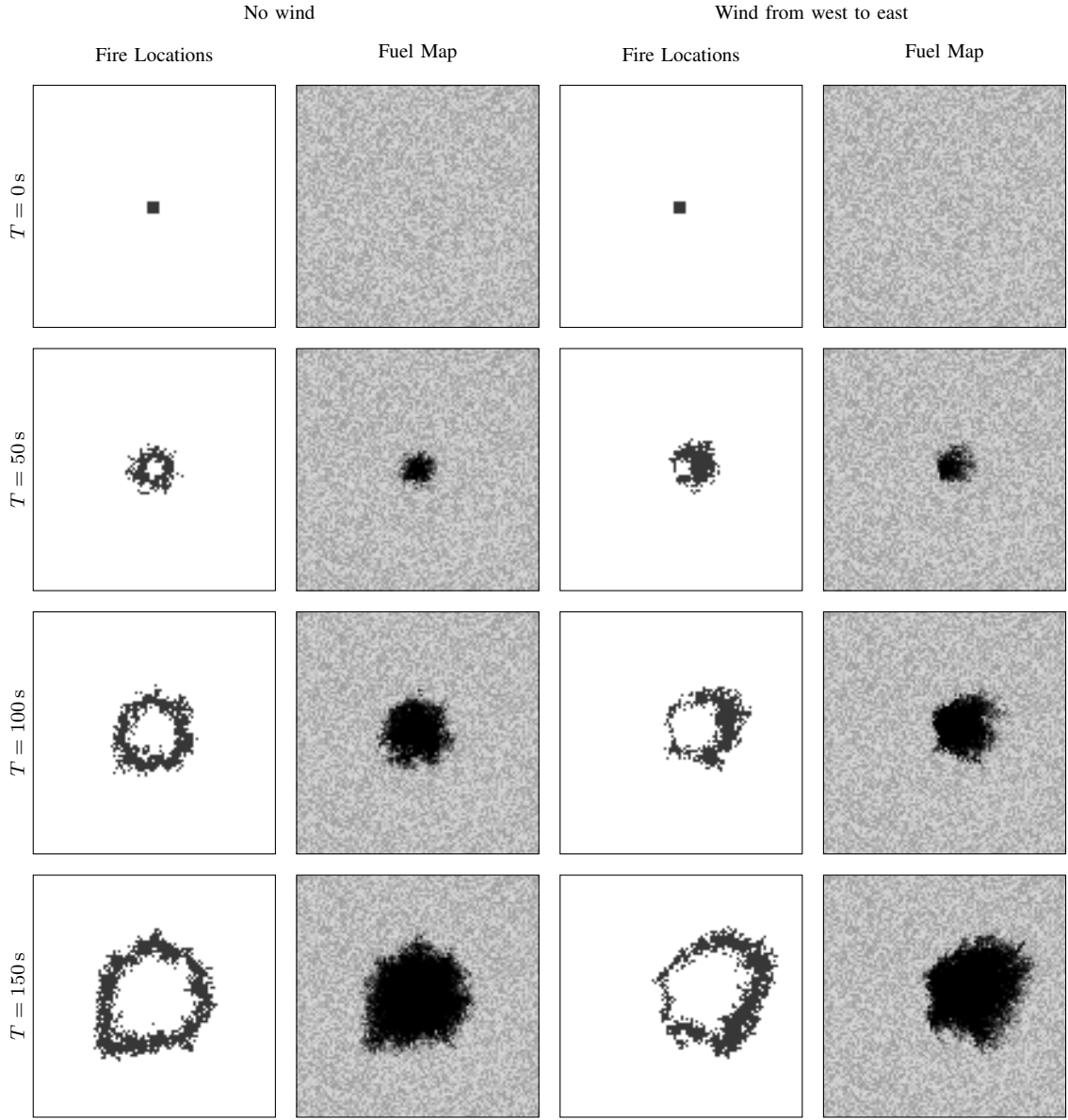

\Cref{fig:FireProp} shows the state of a simulated wildfire for 60 steps in both windless and windy conditions, with each iteration lasting \SI{2.5}{\second}. Each cell begins with a random amount of fuel between 15 and 20 units. The fire propagates stochastically, and while the the wildfire without wind forms a roughly circular and centered ring, the windy wildfire grows more quickly to the east. An effective wildfire monitoring aircraft must be able to track the wildfire growth well under the uncertainty caused by stochasticity and different wind conditions.

\subsection{Aircraft Model}
Assuming that the aircraft maintain steady-level flight at constant speed as they fly around the wildfire the dynamics can be modeled using Dubins' kinematic model \cite{dubins}. Given position $(x,y)$, constant speed $v$ of \SI{20}{\meter\per\second}, and gravitational constant $g$, the state and position are updated using 
\begin{equation} \label{eq_dyn2}
\dot{x} = v\cos\psi,\;  \dot{y}=v\sin\psi,\;  \dot{\psi} = \frac{g\tan\phi}{v}
\end{equation}

Trajectories can be controlled by adjusting the bank angle. In this model, the aircraft has two possible actions at each time step: to increase or decrease the bank angle by 5 deg. By modulating the bank angle by small amounts at a frequency of \SI{10}{\hertz}, precise bank angles and rapid turns can be commanded. The aircraft can track the bank angle commands using a PD controller, allowing the aircraft to follow complex trajectories \cite{julian2017neural}. In simulation the bank angle is assumed to equal the commanded bank angle; experimental flights confirm that an aircraft can track these bank angle commands \cite{julian2017neural}. In order to simplify the action space and enable more efficient policy training, an action to maintain the current bank angle is omitted. Maintaining a bank angle can be done approximately by alternating between increasing and decreasing the bank angle. In addition, the bank angle is capped at a magnitude of 50 deg, so choosing actions to exceed this limit will result in no change to the commanded bank angle.

\subsection{POMDP implementation}
To formulate the problem as a POMDP, the state and action spaces must be defined. The action space is composed of the two possible actions to increase or decrease the bank angle. The state space should capture the necessary features of the aircraft positions and orientations to allow the vehicles to define the scenario geometry. While the global position and orientation of the aircraft are maintained as variables in the simulation, they include redundant information not relevant for making good decisions. Instead of using the true aircraft positions, only the relative position and orientation of the other aircraft is needed. In addition, the bank angles of the two aircraft are necessary to predict the movement of the aircraft. As a result, the POMDP state space includes five continuous state variables:
\begin{enumerate}
	\item $\phi_{0}$: bank angle of ownship
	\item $\rho$: range to other aircraft
	\item $\theta$: bearing angle to other aircraft relative to current heading direction of the ownship
	\item $\psi$: heading angle of other aircraft relative to current heading direction of the ownship
	\item $\phi_{1}$: bank angle of other aircraft
\end{enumerate}

The POMDP approach also requires observation and reward models. The observations can be used with the state variables as a basis for decision making. The next sections present two approaches for modeling limited sensor information and incorporating the observations in the POMDP framework.	

\section{Observation-based Approach \label{sec:Approach1}}
	This section explains a method for making decisions based on immediate wildfire sensor information.
	
	\subsection{Observations \label{sec:Observation1}}
	Due to sensor limitations, each aircraft will not be able to observe the global state of the wildfire. Instead, each aircraft will get an observation of the fire relative to its own location and orientation. The observations are modeled as an image obtained from the true wildfire state given the aircraft's current position and heading direction. The observations are also composed of cells with binary values representing whether a cell is burning.
	
\begin{figure}
		\centering
		\begin{tikzpicture}[]
\begin{groupplot}[height = {6cm},width={6cm}, group style={horizontal sep=2.0cm, vertical sep=1.4cm, group size=1 by 2}]
\nextgroupplot [xlabel={East (m)}, ylabel={North (m)}, enlargelimits = false, axis on top]\addplot [point meta min=0, point meta max=1] graphics [xmin=10, xmax=1000, ymin=10, ymax=1000] {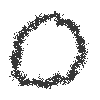};
\node[aircraft top,fill=black,draw=white, minimum width=1cm,rotate=-75.55089,scale = 0.45] at (axis cs:220.99437407148793, 332.578661192556) {};

\nextgroupplot [ylabel = {Crossrange (m)}, xlabel = {Downrange (m)}, enlargelimits = false, axis on top]\addplot [point meta min=0, point meta max=1] graphics [xmin=-500, xmax=500, ymin=-500, ymax=500] {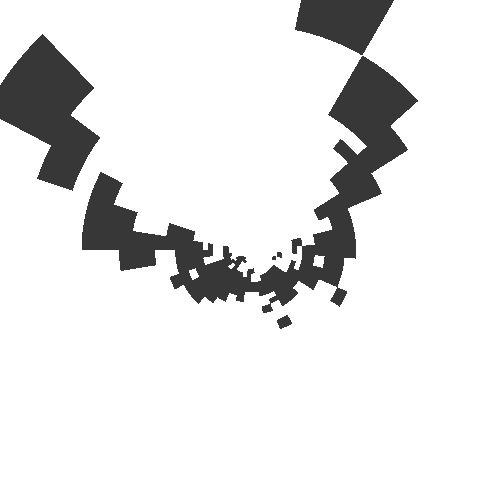};
\node[aircraft top,fill=black,draw=white, minimum width=1cm,rotate=0,scale = 0.45] at (axis cs:0.0, 0.0) {};
\end{groupplot}

\end{tikzpicture}
		\caption{The true wildfire (top) and the aircraft's wildfire observation (bottom)}
		\label{fig:SensorIter}
	\end{figure}

	Since the observations are relative to the aircraft, the observations are polar with 40 range and 30 angular cutpoints.  In this implementation, the interval between range cutpoints begins at \SI{10}{\meter} and ends at \SI{100}{\meter} with a maximum range of \SI{500}{\meter}. Locations closer to the aircraft are perceived with greater resolution than locations further away. This formulation does not attempt to model the error characteristics of on-board sensors such as visual and infrared cameras, or distortions caused by terrain or other factors. The aircraft is assumed to be capable of capturing images of the fire in order to compute the observation, as shown in \cref{fig:SensorIter}.
	
	Each point in the observation image is set to the value of the closest cell in the wildfire map. For cells farther away, this approach could yield inaccurate estimates of the wildfire's boundary. For example, the wildfire front far from the aircraft in \cref{fig:SensorIter} shows gaps in the observation image. In addition, wildfires outside the range of the aircraft will not appear in the observation at all. Thus, the aircraft must make decisions based on imperfect information.

	\subsection{Rewards}
	The aircraft will aim to select actions that maximize accumulated, or, alternatively, minimize accumulated costs. The reward function has four major components:
	\begin{enumerate}
		\item Penalty for distance from fire front
		\item Penalty for non-burning cells nearby
		\item Penalty for high bank angles
		\item Penalty for closeness to the other aircraft
	\end{enumerate}
	 These penalties encourage desired behavior in wildfire monitoring. Since decisions are based on immediate observations, the penalties are approximated from the observations. Penalizing the aircraft for its minimum distance to the fire front, defined as $\min d_s$, promotes a policy where the aircraft quickly flies towards the fire front. By penalizing the aircraft for non-burning cells within some small radius $r_0$ around the aircraft, the aircraft is encouraged to fly over the center of the main fire front. To prevent the aircraft from banking in tight circles over one area of the fire, the bank angle is penalized, making low magnitude bank angles preferable. Lastly, the aircraft observations will be redundant if they cover the same space, so the aircraft are penalized for closeness to each other. This system is not a collision avoidance system, so the aircraft are assumed to operate at different altitude bands to prevent collision. In summary, the penalties given to the aircraft can be written as
	\begin{align}
		r_1 &= -\lambda_1 \min_{\{s \in S \mid B_t(s)\}} d_s \\
		r_2 &= -\lambda_2 \sum_{\{s \in S \mid d_s <r_0\}} 1-B_t(s) \\
		r_3 &= -\lambda_3 \phi_{0}^2 \\
		r_4 &= -\lambda_4 \exp\left(-\frac{\rho}{c}\right)
	\end{align}
	with tuning parameters $\lambda_1$, $\lambda_2$, $\lambda_3$, $\lambda_4$, $r_0$, and $c$. In these equations, $d_s$ is the distance from the aircraft to cell $s$.

\section{Belief-based Approach \label{sec:Approach2}}

\begin{figure*}
	\centering
	\begin{tikzpicture}[]
\begin{groupplot}[height = {5.5cm}, width = {5.5cm}, group style={horizontal sep=0.3cm, vertical sep=0.3cm, group size=3 by 3}]
\nextgroupplot [ylabel = {$T = \SI{0}{\second}$}, title = {Fire Locations}, ticks=none, enlargelimits = false, axis on top]\addplot [point meta min=0, point meta max=1] graphics [xmin=10, xmax=1000, ymin=10, ymax=1000] {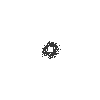};
\node[aircraft top,fill=black,draw=white, minimum width=1cm,rotate=11.459155902,scale = 0.45] at (axis cs:209.80066577841242, 201.98669330795056) {};
\node[aircraft top,fill=black,draw=white, minimum width=1cm,rotate=114.59155902,scale = 0.45] at (axis cs:695.8385316345283, 809.0929742682567) {};

\nextgroupplot [title = {Fire Belief}, ticks=none, enlargelimits = false, axis on top]\addplot [point meta min=0, point meta max=1] graphics [xmin=10, xmax=1000, ymin=10, ymax=1000] {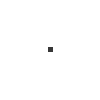};
\node[aircraft top,fill=black,draw=white, minimum width=1cm,rotate=11.459155902,scale = 0.45] at (axis cs:209.80066577841242, 201.98669330795056) {};
\node[aircraft top,fill=black,draw=white, minimum width=1cm,rotate=114.59155902,scale = 0.45] at (axis cs:695.8385316345283, 809.0929742682567) {};

\nextgroupplot [title = {Time Since Last Visited (s)}, ticks=none, enlargelimits = false, axis on top, colormap={wb}{gray(0cm)=(0); gray(1cm)=(1)}, colorbar, colorbar style={ytick={0,50,100,150,200,250}}]
\addplot [point meta min=0, point meta max=256] graphics [xmin=10, xmax=1000, ymin=10, ymax=1000] {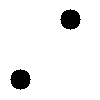};
\node[aircraft top,fill=black,draw=white, minimum width=1cm,rotate=11.459155902,scale = 0.45] at (axis cs:209.80066577841242, 201.98669330795056) {};
\node[aircraft top,fill=black,draw=white, minimum width=1cm,rotate=114.59155902,scale = 0.45] at (axis cs:695.8385316345283, 809.0929742682567) {};

\nextgroupplot [ylabel = {$T = \SI{25}{\second}$}, ticks=none, enlargelimits = false, axis on top]\addplot [point meta min=0, point meta max=1] graphics [xmin=10, xmax=1000, ymin=10, ymax=1000] {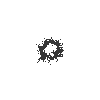};
\node[aircraft top,fill=black,draw=white, minimum width=1cm,rotate=96.3376,scale = 0.45] at (axis cs:532,558) {};
\node[aircraft top,fill=black,draw=white, minimum width=1cm,rotate=-124.27276,scale = 0.45] at (axis cs:378.9179380894684, 566.6777284185058) {};

\nextgroupplot [ticks=none, enlargelimits = false, axis on top]\addplot [point meta min=0, point meta max=1] graphics [xmin=10, xmax=1000, ymin=10, ymax=1000] {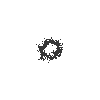};
\node[aircraft top,fill=black,draw=white, minimum width=1cm,rotate=96.3376,scale = 0.45] at (axis cs:532,558) {};
\node[aircraft top,fill=black,draw=white, minimum width=1cm,rotate=-124.27276,scale = 0.45] at (axis cs:378.9179380894684, 566.6777284185058) {};

\nextgroupplot [ticks=none, enlargelimits = false, axis on top, colormap={wb}{gray(0cm)=(0); gray(1cm)=(1)}, colorbar,colorbar style={ytick={0,50,100,150,200,250}}]\addplot [point meta min=0, point meta max=256] graphics [xmin=10, xmax=1000, ymin=10, ymax=1000] {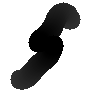};
\node[aircraft top,fill=black,draw=white, minimum width=1cm,rotate=96.3376,scale = 0.45] at (axis cs:532,558) {};
\node[aircraft top,fill=black,draw=white, minimum width=1cm,rotate=-124.27276,scale = 0.45] at (axis cs:378.9179380894684, 566.6777284185058) {};

\nextgroupplot [ylabel = {$T = \SI{100}{\second}$}, ticks=none, enlargelimits = false, axis on top]\addplot [point meta min=0, point meta max=1] graphics [xmin=10, xmax=1000, ymin=10, ymax=1000] {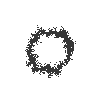};
\node[aircraft top,fill=black,draw=white, minimum width=1cm,rotate=127.48,scale = 0.45] at (axis cs:655.4,597.5514) {};
\node[aircraft top,fill=black,draw=white, minimum width=1cm,rotate=-59.668,scale = 0.45] at (axis cs:312.27,376.974) {};

\nextgroupplot [ticks=none, enlargelimits = false, axis on top]\addplot [point meta min=0, point meta max=1] graphics [xmin=10, xmax=1000, ymin=10, ymax=1000] {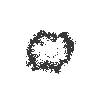};
\node[aircraft top,fill=black,draw=white, minimum width=1cm,rotate=127.48,scale = 0.45] at (axis cs:655.4,597.5514) {};
\node[aircraft top,fill=black,draw=white, minimum width=1cm,rotate=-59.668,scale = 0.45] at (axis cs:312.27,376.974) {};

\nextgroupplot [ticks=none, enlargelimits = false, axis on top, colormap={wb}{gray(0cm)=(0); gray(1cm)=(1)}, colorbar,colorbar style={ytick={0,50,100,150,200,250}}]\addplot [point meta min=0, point meta max=256] graphics [xmin=10, xmax=1000, ymin=10, ymax=1000] {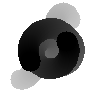};
\node[aircraft top,fill=black,draw=white, minimum width=1cm,rotate=127.48,scale = 0.45] at (axis cs:655.4,597.5514) {};
\node[aircraft top,fill=black,draw=white, minimum width=1cm,rotate=-59.668,scale = 0.45] at (axis cs:312.27,376.974) {};

\end{groupplot}

\end{tikzpicture}
	\caption{Belief maps over time}
	\label{fig:BeliefInit}
\end{figure*}
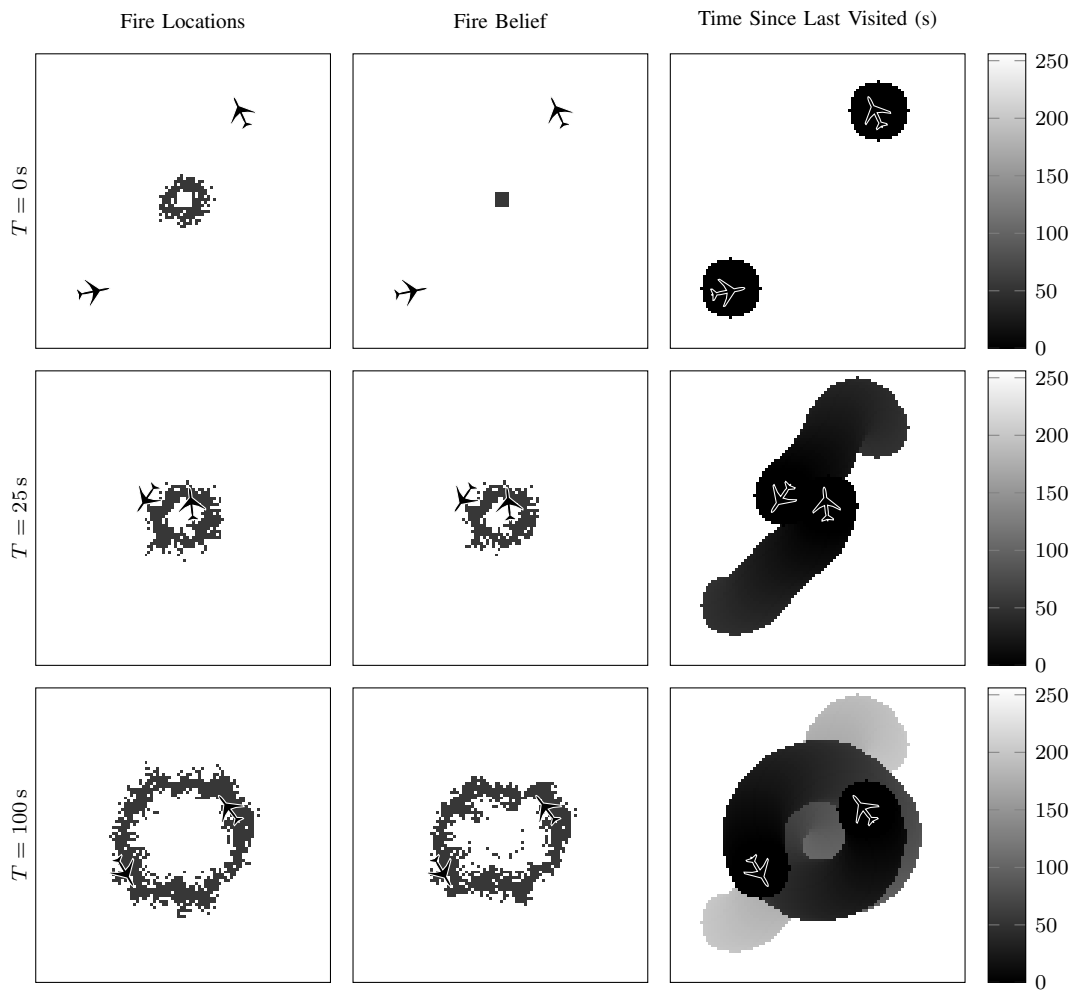

	The observation-based formulation is simple and effective, but it has a few limitations. Decisions are based solely on immediate observations, and that information is discarded for all future decisions. Furthermore, the objective is to minimize the uncertainty of the wildfire's extent at any given time. While the rewards encourage behaviors that lead to good wildfire monitoring trajectories, they are not directly tied to minimizing wildfire uncertainty. The approach presented in this section addresses these issues by using the same solution framework and dynamics. However, this approach bases decisions on an internally maintained belief of the wildfire's location rather than an observation, which allows information to be preserved over time. 
	
	\subsection{Belief \label{sec:Belief}}

	 The belief image is a $100\times100$ grid, the same size as the true map of fire locations. Each cell in the grid has two values: a binary value indicating whether or not the cell is believed to be on fire, and an integer value indicating how many time steps have passed since the cell was visited. As the aircraft flies around the map, each cell within a \SI{100}{\meter} radius of the aircraft is said to be visited, and the values of that cell are updated. For visited cells, the belief that the cell is on fire is set to its true value, and the time elapsed value is set to 0. If a cell is not visited, then the belief that the cell is on fire remains the same, and the time since last visited is incremented by 1 until reaching a maximum of 255 time steps. Additionally, this belief map is shared among and updated by all aircraft, which allows them to collaborate on the wildfire's locations.
	
	The first row of plots in \cref{fig:BeliefInit} depict an initial configuration for a wildfire simulation, where the wildfire has expanded from its initial seed for \SI{30}{\second} before beginning simulation. Although the fire has spread, the belief map is still the initial seed. This map will get updated as the aircraft move around the map. \Cref{fig:BeliefInit} also shows a map of each cell's time elapsed since last visited. Intuitively, the initial map shows values of 0 immediately around each aircraft, and the maximum value of 255 in all other places.
	
	Furthermore, \cref{fig:BeliefInit} shows the belief maps after \SI{100}{\second} have passed. In the example shown here, the aircraft fly toward the center of the map where the initial belief map shows fire. The aircraft begin to spiral along the wildfire boundary, continually updating the fire belief map as they fly. As shown in the fire belief map, the fire boundary where the aircraft recently visited are accurate, but the boundary locations not recently visited may be outdated and different from the true boundary location.
	
	The belief map can be treated as a $100\times100$ image with two channels. To help the algorithm learn better policies, as discussed in \cref{sec:Solution}, this belief image is centered and aligned with the ownship before being used. So although all aircraft share the same belief map, the image that is used in the learning algorithm will be translated and rotated for each aircraft. The belief image is $100\times100$ as well, and cells that appear on the belief image that are outside the bounds of the original belief map are believed to have no fire and have the maximum time since last visited. Lastly, the values of the time elapsed channel of the belief image are normalized to have a maximum value of one in order to make the range of values of the two image channels equal.

	\subsection{Rewards}
	Using the belief state formulation, the rewards can be formulated to minimize the true objective. The aircraft is rewarded when cells in the belief map that were thought to be non-burning are visited and seen to be burning. This reward encourages the aircraft to fly along the fire front and observe as much of the wildfire's spread as possible. The aircraft is not rewarded for visiting a cell believed to be burning and observing that it is not, since these cells are behind the fire front and are assumed to extinguish eventually. If viewing these cells is also valuable, then the reward function can be easily modified to reward visiting these cells too. Furthermore, the aircraft should cooperate to minimize wildfire uncertainty, so each aircraft is rewarded for all updated burning cells, not just the ones that it visited. This encourages the aircraft to collaborate rather than fighting to view new areas of the fire front.
	
	Although these rewards work well when using two aircraft, experimental results in \cref{sec:Results} show that the aircraft do not maintain separation when additional aircraft are added. As a result, a small penalty for proximity to the other aircraft is added, which helps the controller to prioritize aircraft separation even in multi-aircraft scenarios.
	
\section{Solution Method \label{sec:Solution}}
	The solution to a POMDP is a policy that maps observations to actions. Because the space of possible wildfire observations is large, an approximate method must be used. Deep reinforcement learning was used to obtain approximately optimal control strategies for the two formulations. Deep reinforcement learning (DRL) involves training a neural network to represent a policy mapping states or observations to actions that maximizes the expected accumulated reward. The network is trained offline through simulation, and the trained network can be used on-board the aircraft to continuously map observations to actions. 
\subsection{Deep Reinforcement Learning \label{sec:DRL}}
	In deep reinforcement learning, each state-action pair has a value, $Q(s,a)$, which represents the expected value of taking action $a$ from state $s$. The optimal Q-values should follow the Bellman equation
	\begin{equation}
		Q(s,a) = r(s) + \gamma\sum_{s' \in S} p(s' \mid s,a) \max_{a' \in A} Q(s',a')
	\end{equation}
	where $\gamma$ is the discount factor, $r(s)$ is the reward received for being in state $s$, and $p(s' \mid s,a)$ is the probability of transitioning to state $s'$ by taking action $a$ from state $s$. The discount factor, set to $0.99$ here, is used to discount future rewards and help the algorithm converge. Given a set of Q-values the policy can be computed as
	\begin{equation}
	\pi(s) = \argmax_{a \in A} Q(s,a)
	\end{equation}
	 However, computing the Q-values for each state-action pair is intractable for such a large state space. Furthermore, the transition probabilities $p(s' \mid s,a)$ are unknown. Instead, the Q-values are estimated using function approximators trained through interaction with a simulation environment. Deep neural networks are capable of representing complex functions, and their parameters are easily and quickly optimized through gradient descent. Simulations are run to generate tuples of states, actions, rewards, and next states, which can be used to compute the Bellman error, defined as 
	\begin{equation}
		E = r + \gamma \max_{a' \in A} Q(s',a') - Q(s,a)
		\label{eq:bell}
	\end{equation}
	
	The Bellman error can be minimized through mean squared error loss and gradient descent methods. When using function approximation, updating the parameters to change the $Q$ values of one state will also update the $Q$ values of very similar states, which can lead to faster convergence. However, this ability can lead to unwanted side effects. Since a given state $s$ and next state $s'$ will be similar, changing $Q(s,a)$ will also change $Q(s',a')$, which can increase the error and render the learning process unstable.
	
	There are techniques to mitigate such instabilities \cite{mnih2015human}. With experience replay, tuples of state, action, reward, and next state can be stored in a large repository, and random samples can be drawn to train the network in order to split up trajectories where states are similar. Before network training begins, this repository is supplied with a substantial number of samples to ensure that the network does not over-fit to a few trajectories.
	
	In addition, $Q(s',a')$ is computed from a fixed target network, which is updated periodically, ensuring that $Q(s,a)$ is targeting a fixed value rather than chasing a moving target and growing unstably. The network must be trained for a long enough period before the target network is updated to ensure that training loss does not become unstable.
	
	Finally, deep Q-learning must decide between exploration and exploitation when selecting actions. Exploring by choosing random actions when generating trajectories can reveal good strategies that may not receive higher rewards initially. However, exploiting by taking actions that previously yielded high reward can refine strategies and greatly improve performance. This trade-off is addressed by using an $\epsilon$-greedy action selection strategy. In this strategy, a random action is selected with probability $\epsilon$ while the action dictated by the policy is followed with probability $1-\epsilon$. Initially $\epsilon=1$, but this parameter decreases linearly to 0.1 over time \cite{mnih2015human}. This exploration strategy means that the learner will explore many different trajectories initially and then fine tune the optimal policy as the $Q$ values converge. 
	
	If more than two aircraft are desired, the same neural network controller can be used. Since the network estimates the score of each action considering pairwise interactions, the action scores can be computed by summing the scores of each pair-wise interaction \cite{Chryssanthacopoulos2012decomp}. If there are $n$ additional aircraft with states $s_1,\dots ,s_n$, the best action to take for the aircraft is 
	\begin{equation}
	\pi(s) = \argmax_{a \in A} \sum_{j=1}^{n} Q(s_j,a)
	\end{equation}
	
	DRL generally maps states to actions, but in this formulation only observations of the full wildfire are available. Therefore, the state used by this DRL algorithm is composed of the continuous aircraft state variables as well as either the observation or belief image, depending on which approach is used.

\subsection{Network Architecture \label{sec:Network}}
 
	Both approaches presented here use the same network architecture, although the inputs to the network have different dimensions. Unlike network architectures for playing video games, which use only image information as an input, the network input for the wildfire problem includes an image as well as continuous inputs \cite{mnih2015human}. Therefore, the neural network begins with two separate networks: a convolutional network for the image, and fully connected layers for the continuous state variables. Convolutional neural networks exploit similarities between local regions of pixels, allowing them to be efficient and effective with high-dimensional inputs. The convolutional neural network contains max pooling layers. These layers take the maximum of each non-overlapping $2\times2$ region in the previous layer, which significantly reduces the amount of weights needed for future layers. This helps to keep the network to a manageable size without significant change in performance. The convolutional and max pooling layers are then flattened and connected to two more fully connected layers, ending with 100 units in the last layer.
	
\begin{figure}
	\centering
	\tikzstyle{dense} = [rectangle, draw, fill=black!5, node distance =1cm, text width = 10em, rounded corners,text centered, thick,style={font=\footnotesize}]
\tikzstyle{dense2} = [rectangle, draw, fill=black!5, node distance =1cm, text width = 15em, rounded corners,text centered, thick,style={font=\footnotesize}]

\tikzstyle{conv} = [rectangle, draw, fill=black!5, node distance=2cm, text width=10em, text centered, rounded corners, thick,style={font=\footnotesize}]
\tikzstyle{max} = [rectangle, draw, fill=black!5, node distance=2cm, text width=10em, text centered, rounded corners, thick,style={font=\footnotesize}]
\tikzstyle{l} = [draw, -latex',thick]

\begin{tikzpicture}

	\def\vertDense{1.395cm}
    \node [dense] (d1) {Dense Layer: 100};
    \node [dense, below of=d1,node distance = \vertDense] (d2) {Dense Layer: 100};
    \node [dense, below of=d2,node distance = \vertDense] (d3) {Dense Layer: 100};
    \node [dense, below of=d3,node distance = \vertDense] (d4) {Dense Layer: 100};
    \node [dense, below of=d4,node distance = \vertDense] (d5) {Dense Layer: 100};
    
    \node [conv, right of=d1, node distance = 4.7cm] (c1) {Conv. Layer: 64 3x3};
    \node [max, below of=c1, node distance = 0.8cm]  (m1) {Max Pool Layer: 2x2};
	\node [conv, below of=m1, node distance = 0.8cm] (c2) {Conv. Layer: 64 3x3};
	\node [max, below of=c2, node distance = 0.8cm]  (m2) {Max Pool Layer: 2x2};
	\node [conv, below of=m2, node distance = 0.8cm] (c3) {Conv. Layer: 64 3x3};
	\node [max, below of=c3, node distance = 0.8cm]  (m3) {Max Pool Layer: 2x2};
	\node [dense, below of=m3, node distance = 0.8cm] (d6) {Dense Layer: 500};
	\node [dense, below of=d6, node distance = 0.8cm] (d7) {Dense Layer: 100};
	
	\node [dense2, below of=d5, xshift=2.35cm,node distance=1.1cm] (d8) {Dense Layer: 200};
	\node [dense2, below of=d8, node distance=1.1cm] (d9) {Dense Layer: 200};
	\node [dense, below of=d9, node distance=1.1cm] (d10) {Output Layer: 2};
	
	\path [l] (d1) edge[out=270, in=90] (d2);
	\path [l] (d2) edge[out=270, in=90] (d3);
	\path [l] (d3) edge[out=270, in=90] (d4);
	\path [l] (d4) edge[out=270, in=90] (d5);
	\path [l] (d5) edge[out=270, in=90] (d8);
	\path [l] (d8) edge[out=270, in=90] (d9);
	\path [l] (d9) edge[out=270, in=90] (d10);
	
	\path [l] (c1) edge[out=270, in=90] (m1);
	\path [l] (m1) edge[out=270, in=90] (c2);
	\path [l] (c2) edge[out=270, in=90] (m2);
	\path [l] (m2) edge[out=270, in=90] (c3);
	\path [l] (c3) edge[out=270, in=90] (m3);
	\path [l] (m3) edge[out=270, in=90] (d6);
	\path [l] (d6) edge[out=270, in=90] (d7);
	\path [l] (d7) edge[out=270, in=90] (d8);
	
	\node [below of=d10,xshift=-1.0cm,node distance=0.15cm] (o1_beg)  {};
	\node [below of=d10,xshift=-1.0cm,node distance=1.1cm] (o1_end)  {};
	\path [l] (o1_beg) edge[out=270, in=90] (o1_end);
	\node [below of=o1_end,node distance=0.2cm] (Q1) {$Q(s,a_1)$};
	
	\node [below of=d10,xshift=1.0cm,node distance=0.15cm] (o2_beg)  {};
	\node [below of=d10,xshift=1.0cm,node distance=1.1cm] (o2_end)  {};
	\path [l] (o2_beg) edge[out=270, in=90] (o2_end);	
	\node [below of=o2_end,node distance=0.2cm] (Q2) {$Q(s,a_2)$};

	\node [above of=d1,xshift=-1.4cm,node distance=0.15cm] (i1_beg)  {};
	\node [above of=d1,xshift=-1.4cm,node distance=1.1cm] (i1_end)  {};
	\path [l] (i1_end) edge[out=270, in=90] (i1_beg);
	\node [above of=i1_end,node distance=0.2cm] (I1) {$\phi_{0}$};
	
	\node [above of=d1,xshift=1.4cm,node distance=0.15cm] (i2_beg)  {};
	\node [above of=d1,xshift=1.4cm,node distance=1.1cm] (i2_end)  {};
	\path [l] (i2_end) edge[out=270, in=90] (i2_beg);	
	\node [above of=i2_end,node distance=0.2cm] (I2) {$\phi_{1}$};
	
	\node [above of=d1,xshift=-0.7cm,node distance=0.15cm] (i3_beg)  {};
	\node [above of=d1,xshift=-0.7cm,node distance=1.1cm] (i3_end)  {};
	\path [l] (i3_end) edge[out=270, in=90] (i3_beg);
	\node [above of=i3_end,node distance=0.2cm] (I3) {$\rho$};
	
	\node [above of=d1,xshift=0.7cm,node distance=0.15cm] (i4_beg)  {};
	\node [above of=d1,xshift=0.7cm,node distance=1.1cm] (i4_end)  {};
	\path [l] (i4_end) edge[out=270, in=90] (i4_beg);	
	\node [above of=i4_end,node distance=0.2cm] (I4) {$\psi$};
	
	\node [above of=d1,node distance=0.15cm] (i5_beg)  {};
	\node [above of=d1,node distance=1.1cm] (i5_end)  {};
	\path [l] (i5_end) edge[out=270, in=90] (i5_beg);
	\node [above of=i5_end,node distance=0.2cm] (I5) {$\theta$};
	
	\node [above of=c1,node distance=0.15cm] (i6_beg)  {};
	\node [above of=c1,node distance=1.1cm] (i6_end)  {};
	\path [l] (i6_end) edge[out=270, in=90] (i6_beg);
	\node [above of=i6_end,node distance=0.15cm] (I6) {Observation/Belief  Image};

\end{tikzpicture}
	\caption{Network architecture}
	\label{fig:NetArch}
\end{figure}
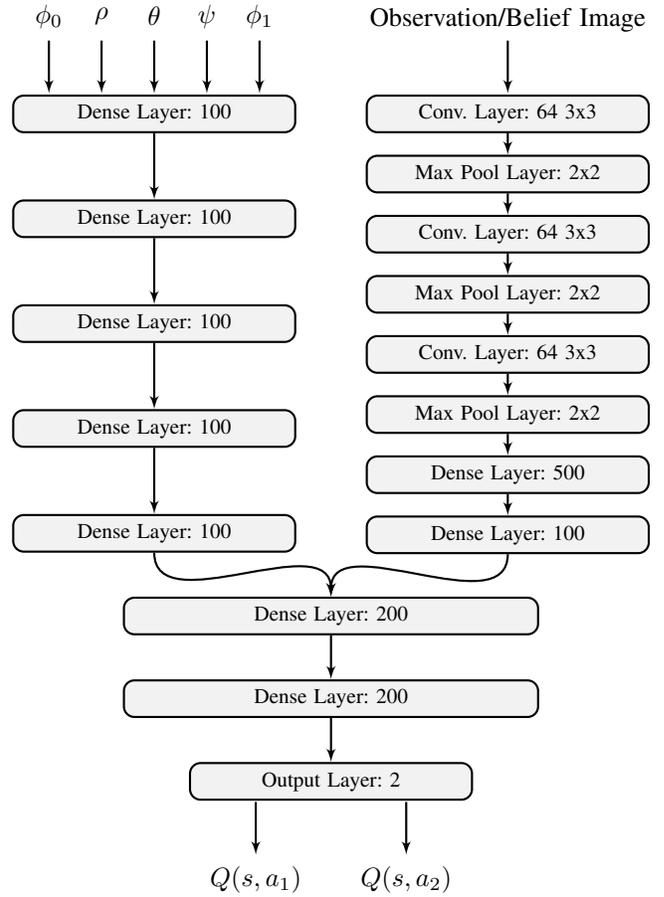

	The fully connected layers for the continuous inputs consist of five hidden layers of 100 units each. At this point, both sides of the network have 100 units each. These two hidden layers are concatenated into a layer of 200 hidden units. The merged layers are followed by two more fully connected hidden layers of 200 units each, which then end with an output layer with two outputs. This network architecture is depicted in \cref{fig:NetArch}.
	
	Each hidden layer in the network uses rectified linear unit activations, which allow positive inputs to pass through unchanged while negative inputs are output as zero. The network parameters are updated using AdaMax optimization, an adaptive stochastic gradient descent based on the infinity norm, which estimates low order moments to change learning parameters and minimize loss very quickly \cite{Adam}. The network is trained on mini-batches of 64 samples drawn from the experience replay. The fixed target network is updated every 1000 training iterations, which allows the loss to converge before moving the target. Training requires 12 hours to finish exploration, and another 12 hours of training further improves network performance.

\section{Baseline Controller \label{sec:Baseline}}
	To better illustrate the performance of the deep neural network approach, a baseline receding horizon controller was created. The receding horizon controller is an online method that uses the same bank angle modifying actions selected at \SI{10}{\hertz}. A trajectory is optimized for $T$ steps assuming the wildfire remains fixed. Then, the aircraft executes the first $\tau$ steps, with $\tau < T$, after which the controller optimizes a new trajectory. Executing only a part of the optimized trajectory allows the aircraft to re-plan as the wildfire propagates. The trajectory is optimized to maximize the observation-based rewards that encourage following along the wildfire front. In addition, the trajectory is planned using the true wildfire map rather than an observation or belief map, which creates a more conservative and accurate baseline controller.
	
	The number of possible trajectories grows exponentially with the number of time steps $T$, though larger $T$ values lead to better performance. Rather than evaluating each possible trajectory, a coordinate descent approach with random restarts was used. Beginning with a random trajectory parameterized by the actions taken at each time step, each action is evaluated to determine if changing the action leads to better performance. This process repeats until no change improves performance, resulting in a locally optimal trajectory. This process is repeated for many random restarts, and the best trajectory is used. Although this method does not guarantee the optimal trajectory is found, the computation time is reduced and allows larger $T$ to be tractable. Simulated experiments revealed that slightly suboptimal trajectories planned over longer horizons produce better results than optimal trajectories planned over shorter horizons. Simultaneously optimizing the trajectories of multiple aircraft was also explored, but since the number of possible joint actions grows exponentially with the number of aircraft, the number of possible trajectories grows large very quickly, and trajectories become difficult to optimize. Experiments showed that optimizing trajectories of aircraft separately led to better performance in a given amount of computation time, so joint trajectories were not used in the baseline controller.

\section{Simulation Results \label{sec:Results}}
	This section summarizes a variety of simulation results. First the performance of the two approaches at different points during training is evaluated for two aircraft. Next the effect of adding more aircraft is investigated, and finally the approaches' generalization to other wildfire shapes is explored.

\subsection{Two aircraft with nominal wildfire seed}
	
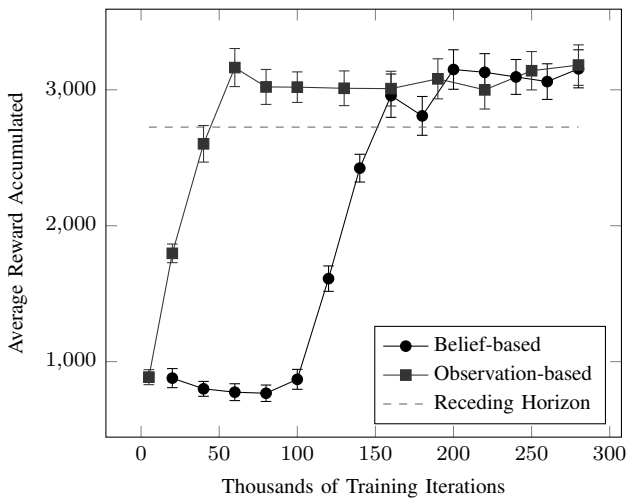
\begin{figure}
		\centering
		\begin{tikzpicture}[]
\begin{axis}[legend pos = {south east}, xlabel = {Thousands of Training Iterations}, ylabel = {Average Reward Accumulated}]
\addplot+ [black,mark options={black}
, error bars/.cd, 
x dir=both, x explicit, y dir=both, y explicit]
table [
x error plus=ex+, x error minus=ex-, y error plus=ey+, y error minus=ey-
] {
x y ex+ ex- ey+ ey-
20.0 878.9   0.0 0.0 70.2 70.2
40.0 800.1   0.0 0.0 54.8 54.8
60.0 775.5   0.0 0.0 62.0 62.0
80.0 767.6   0.0 0.0 60.5 60.5
100.0 870.0  0.0 0.0 73.3 73.3
120.0 1610.8 0.0 0.0 93.5 93.5
140.0 2423.6 0.0 0.0 102.2 102.2
160.0 2956.8 0.0 0.0 159.4 159.4
180.0 2808.1 0.0 0.0 142.8 142.8
200.0 3149.3 0.0 0.0 145.4 145.4
220.0 3128.6 0.0 0.0 137.4 137.4
240.0 3094.6 0.0 0.0 128.3 128.3
260.0 3060.3 0.0 0.0 131.1 131.1
280.0 3154.3 0.0 0.0 139.9 139.9
};
\addlegendentry{Belief-based}
\addplot+ [black!80,mark options={black!80}
, error bars/.cd, 
x dir=both, x explicit, y dir=both, y explicit]
table [
x error plus=ex+, x error minus=ex-, y error plus=ey+, y error minus=ey-
] {
x y ex+ ex- ey+ ey-
5.0 885.9 0.0 0.0 55.064 55.064
20.0 1796.85 0.0 0.0  68.6321 68.6321
40.0 2602.7  0.0 0.0 134.511 134.511
60.0 3163.5  0.0 0.0 140.008 140.008
80.0 3021.05 0.0 0.0 128.573 128.573
100.0 3019.4 0.0 0.0 112.308 112.308
130.0 3011.1 0.0 0.0 127.597 127.597
160.0 3008.4 0.0 0.0 127.87 127.87
190.0 3080.8 0.0 0.0 147.311 147.311
220.0 2999.6 0.0 0.0 141.049 141.049
250.0 3140.25 0.0 0.0 141.034 141.034
280.0 3181.7  0.0 0.0 148.872 148.872
};
\addlegendentry{Observation-based}

\addplot+ [no marks, black!60, dashed]coordinates{
(5.0, 2725.6)
(280.0, 2725.6)
};
\addlegendentry{Receding Horizon}

\end{axis}

\end{tikzpicture}
		\caption{Simulation performance during training}
		\label{fig:PerfPlot}
	\end{figure}

	To understand how the two approaches improve with training, 20 simulations were conducted at different points during training. The reward accumulation of the belief-based network was used as the performance metric because maximizing this quantity is equivalent to minimizing uncertainty in the wildfire's location. The average accumulated reward and standard error for different points during training are shown in \cref{fig:PerfPlot}. The neural network controllers initially perform poorly, but through training the performance improves quickly. The observation-based approach learns more quickly because its reward function incentivizes flying towards the wildfire, while the belief-based approach is only rewarded once the wildfire is near the aircraft. Eventually, the belief-based approach outperforms the observation-based approach by a small amount, which is expected as the belief-based approach was designed to maximize this performance metric, while the observation-based approach optimizes a different reward function.
	
\begin{figure*}
		
		\begin{tikzpicture}
		\matrix [anchor=base] {
			\node {\quad \qquad \footnotesize No wind\qquad\qquad \qquad \qquad \qquad}; & \node {\quad \qquad \qquad \qquad \qquad \footnotesize Wind from west to east};\\
		};
		\end{tikzpicture}
		\centering
		\input{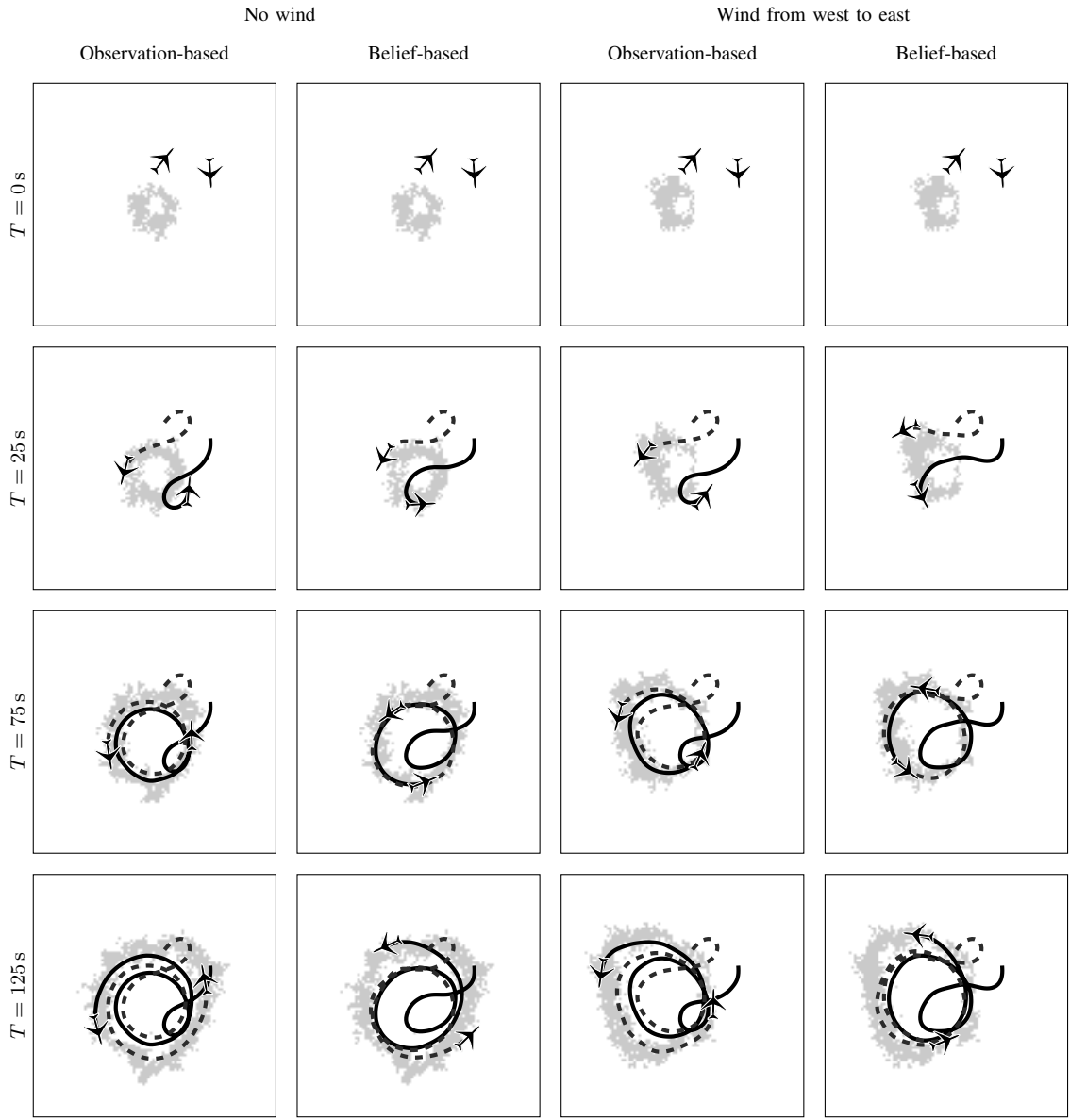}
		\caption{Trajectories of two aircraft using belief and observation-based controllers}
		\label{fig:Paths2}
	\end{figure*}

	After fully training the networks, the decisions made by the neural networks can be understood qualitatively through simulation and inspection of the flight trajectories. The paths taken by two aircraft using both belief and observation-based neural network controllers are plotted with the true wildfire map with and without wind in \cref{fig:Paths2}. The aircraft begin at the same random location for all simulations, and the wildfire is seeded and allowed to grow for \SI{30}{\second} before the aircraft begin to move. 
	
	\Cref{fig:Paths2} illustrate some key features of the two network controllers. After 25 seconds have elapsed, the networks have guided the aircraft to opposite sides of the fire front. Once they reach the fire front, the aircraft are heading in the same direction, so the solid trajectory aircraft turns around. This allows the two aircraft to fly around the fire front and remain far apart from each other. Although tight turns incur penalties for the observation-based network, the network has learned that performing this maneuver early on ensures that the aircraft can gain as much separation on the wildfire as possible, which will be beneficial over time, and the belief-based approach learns the same maneuver. This behavior demonstrates that the network has learned complex strategies beyond greedy actions, and it demonstrates that the aircraft are able to perform coordinated trajectories even though there is no explicit communication. Because both aircraft are trained using the same network for guidance, the aircraft learn to anticipate the movement of the other aircraft, allowing them to perform efficiently and cooperatively.
	
	The belief-based network shows different tendencies than the observation-network, especially early in the simulation. After 25 seconds have elapsed, the aircraft have visited the entire wildfire as well as areas immediately around the fire front. This differs from the observation-based method, which justs orients the aircraft on the fire front without exploring areas near the front. This difference in behavior stems from the differences in received observations. The observation-based approach receives coarse information from a larger radius while the belief-based approach receives perfect information only immediately near the aircraft. As a result, the belief-based approach needs to explore the fire front more to ensure that the wildfire belief is accurate in all areas around the wildfire. After the area is explored, the belief-based approach guides the aircraft around the wildfire front as with the observation-based approach.
	
	Furthermore, the observation-based trajectories are more adaptable to wind. Without wind the trajectories are evenly spaced out as the wildfire grows uniformly in all directions. With wind the trajectories have little separation in areas where wildfire growth is slow and large separation where growth is fast. However, the belief-based approach trajectories are similar in windless and winy conditions, which shows that the belief-based approach does not adapt the trajectories as well in wind. The ability to adapt trajectories will be explored further in \cref{sec_diff_shapes}.
	
	\begin{figure}
		\centering
		\input{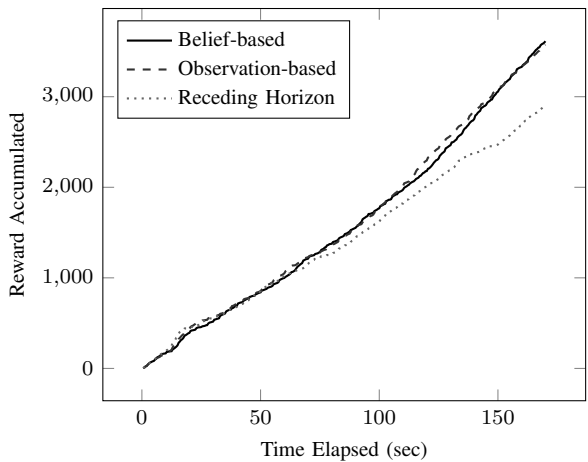}
		\caption{Accumulation of reward during simulation}
		\label{fig:Accum}
	\end{figure}

\begin{figure*}
		\centering
		\input{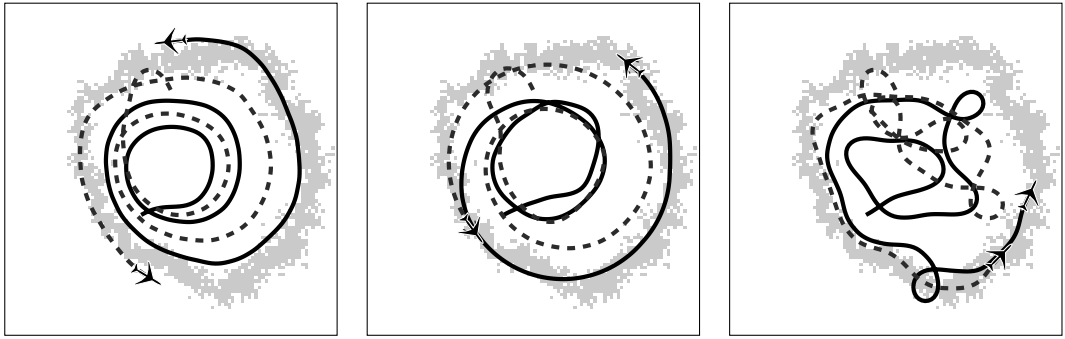}
		\caption{Trajectories during reward accumulation for the observation-based approach (left), belief-based approach (middle), and receding horizon baseline (c)}
		\label{fig:RewardAccumTrajectories}
	\end{figure*}

	Another way to investigate the performance of the two approaches is to compare the accumulation of reward in a single simulation as a function of time. As seen in \cref{fig:Accum}, the belief-based and observation-based methods perform very similarly, but both outperform a receding horizon controller. \Cref{fig:RewardAccumTrajectories} shows the trajectories that generated the reward accumulation plot in \cref{fig:Accum}. The receding horizon controller's trajectories have sharper turns as it tracks the wildfire front, which is due to choosing a trajectory that is only good in the near term and having to correct for the choice in the future. The receding horizon controller is not coordinated, so the two aircraft fly near each other and accumulate less reward. Both network controllers outperform the online receding horizon controller, even though they make decisions with limited wildfire information.
	

\subsection{More than two aircraft}

\begin{figure}
	\centering
	\input{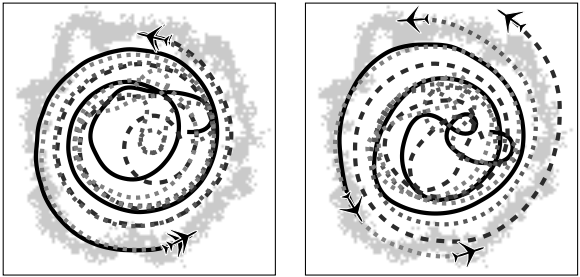}
	\caption{Paths taken by three aircraft using the belief-based approach without the proximity penalty (left) and with the proximity (right)}
	\label{fig:Paths4_penalty}
\end{figure}

As mentioned in \cref{sec:Approach2}, the belief-based approach does not separate aircraft well when additional aircraft are added. Because the observation-based method has an explicit penalty for proximity to other aircraft, the neural network learns a stronger relationship between aircraft distance and $Q$ values. However, without this penalty, the belief-based approach does not learn this relationship well. By adding a small penalty for proximity to other aircraft, the belief-based controller can be trained to separate from other aircraft in multi-aircraft scenarios. \Cref{fig:Paths4_penalty} shows the trajectories generated for four aircraft when the network is trained with and without the proximity penalty. Without the penalty the aircraft fly in two pairs, but with the penalty the aircraft maintain good separation throughout the trajectory.

\begin{figure*}
	\begin{tikzpicture}
	\matrix [anchor=base] {
		\node { \footnotesize No Wind \quad \qquad\qquad \qquad \qquad \qquad}; & \node {\quad \qquad \qquad \qquad \qquad \qquad \footnotesize With Wind \qquad};\\
	};
	\end{tikzpicture}
	\centering
	\input{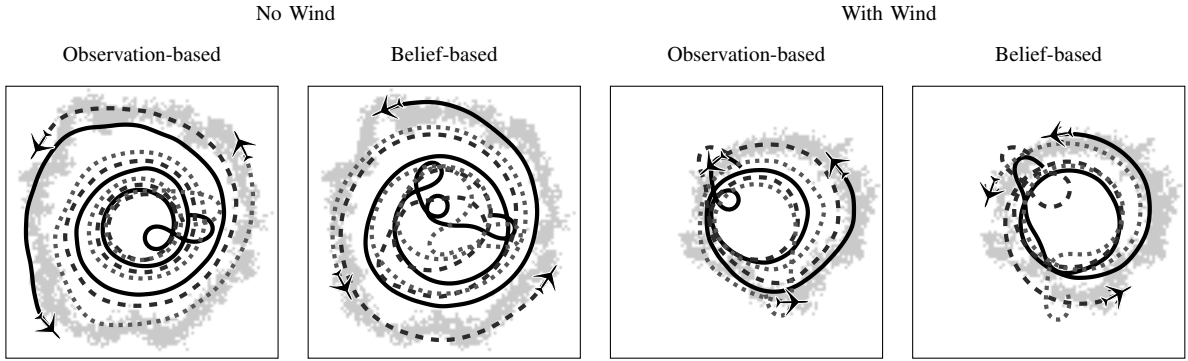}
	\caption{Paths taken by three aircraft}
	\label{fig:Paths3}
\end{figure*}
\begin{figure*}
	\begin{tikzpicture}
	\matrix [anchor=base] {
		\node {\footnotesize No Wind  \quad \qquad\qquad \qquad \qquad \qquad}; & \node {\quad \qquad \qquad \qquad \qquad \qquad \footnotesize With Wind \qquad};\\
	};
	\end{tikzpicture}
	\centering
	\input{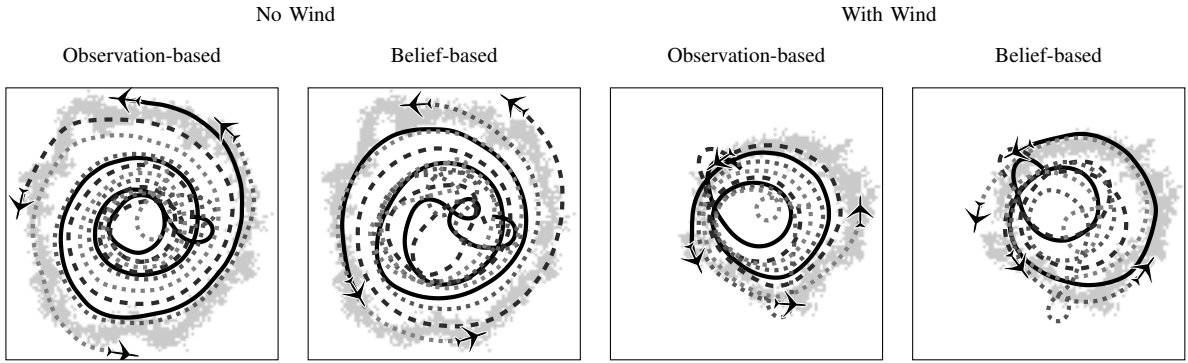}
	\caption{Paths taken by four aircraft}
	\label{fig:Paths4}
\end{figure*}

\Cref{fig:Paths3} shows the trajectories of three aircraft during simulation using both approaches. Both approaches first guide the aircraft to the fire such that they are distributed along the fire front, and then they direct the aircraft outward as they track the growing fire. \Cref{fig:Paths3} shows the final configuration of the aircraft along with the the final shape of the wildfire. In both windless and windy conditions, the aircraft track the fire front while maintaining good separation. Trajectories using four aircraft are depicted in \cref{fig:Paths4}, and again both approaches spread the aircraft around the fire front and track the wildfire well.

The average reward accumulated by the two approaches can be computed to compare how scaling to more aircraft affects their ability to monitor the wildfire. As in \cref{fig:PerfPlot}, the reward from the belief-based approach is used for measuring the performance of all methods. \cref{tableMutliple} show that the belief-based controller slightly outperforms the observation-based controller when using only two aircraft. However, the observation-based method outperforms the belief-based method for more than two aircraft. Even with the separation penalty added to the belief-based reward during training, the observation-based method maintains better separation between aircraft and monitors more of the wildfire.

\begin{table}
	\caption{Average Performances for Different Numbers of Aircraft}\label{tableMutliple}
	\centering
	\begin{tabular}{lcc}
		\toprule
		Num. Aircraft & Observation-based & Belief-based \\
		\midrule
		2 & $3140.5\pm130.1$  & $3154.3\pm139.9$  \\
		3 & $3390.3\pm146.2$  & $3337.2\pm161.8$  \\
		4 & $3546.9\pm152.1$  & $3463.8\pm164.8$  \\
		\bottomrule
	\end{tabular}
\end{table}

\subsection{Different wildfire shapes}\label{sec_diff_shapes}

In real scenarios, the wildfire may not always be circular in shape. For the neural network guidance system to be effective, it must be able to adapt to different wildfire shapes. Since the networks were only trained on roughly circular shapes, this investigation is also a measure of the networks' ability to generalize to new situations unseen during training.

\begin{figure*}
	\centering
	\input{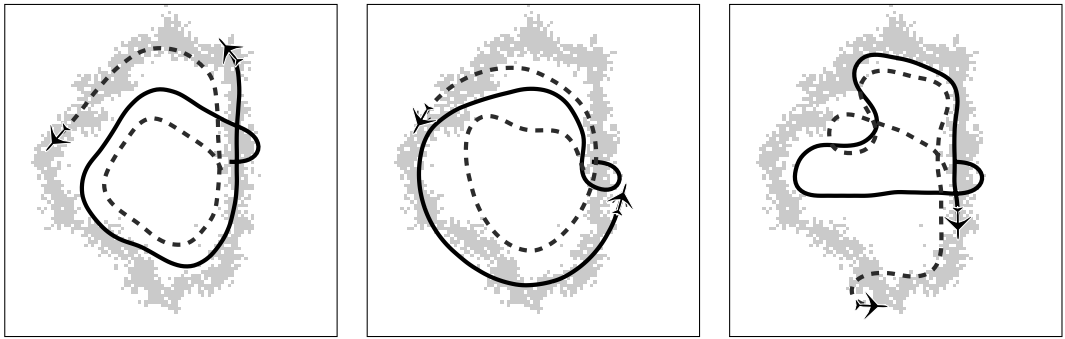}
	\caption{Paths taken by aircraft following a T-shaped fire front using the observation-based approach (left), belief-based approach (middle), and receding horizon baseline (right)}
	\label{fig:PathsCross}
\end{figure*}

The first test shape seeds the wildfire with a T-shaped pattern by initializing the wildfire to extend from the center in western, northern, and southern directions. This pattern creates a wildfire boundary with straight, convex, and concave sections, which is more complex than a simple ring of wildfire. The flight paths and final configurations for the two approaches are plotted in \cref{fig:PathsCross}. Visually, the observation-based approach performs better here, as this method is designed to follow wherever the wildfire front leads. The aircraft are able to fly along straight sections well and generally track the fire front, although spacing is not maintained as effectively. The belief-based approach does not follow the wildfire front as closely and instead chooses to fly in a roughly circular shape. As seen in \cref{tableShapes}, the average reward accumulated on wildfires with this shape is higher when using the observation-based approach. Note that \cref{tableShapes} is not a performance comparison between the different wildfire shapes, since different shapes are simulated for different amounts of time and with different amounts of wildfire present. Therefore, the values will differ between shapes.

\begin{figure*}
	\centering
	\input{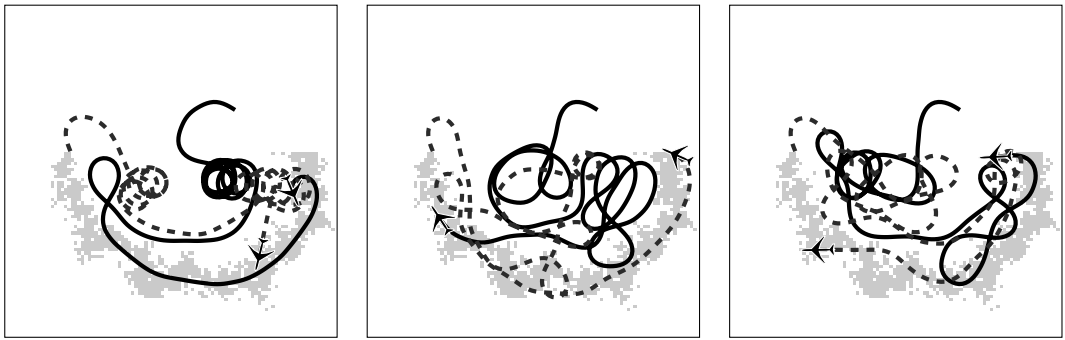}
	\caption{Paths taken by aircraft following an arc-shaped fire front using the observation-based approach (left), belief-based approach (middle), and receding horizon baseline (right)}
	\label{fig:PathsArc}
\end{figure*}

The second shape tested seeds the wildfire nominally, but the cells in the upper half of the land area are devoid of fuel. This situation represents a case where the fire cannot extend past a boundary, perhaps due to some body of water or road. The resulting shape is an arc, which presents new challenges for the aircraft because the wildfire boundary is no longer closed. In initial experiments, the neural network controllers performed poorly on open shapes, so the final controllers were trained with both open and closed wildfire shapes. As seen in \cref{fig:PathsArc}, the behavior of the aircraft changes when the shape is open. The observation-based approach prioritizes aircraft separation, but separation is difficult to maintain when the contour is open. As a result, the aircraft tend to occupy different end-points of the arc and occasionally fly along the boundary. This aversion to flying near other aircraft limits the ability of the observation-based approach to monitor the wildfire. In comparison, the belief-based approach performs much better, as seen in \cref{tableShapes}. The aircraft fly back and forth along the boundary exploring regions of the fire front that have not been visited recently, allowing the aircraft to monitor the wildfire well. 

\begin{table}
	\caption{Average Performance for Different Wildfire Shapes \label{tableShapes}}
	\centering
	\begin{tabular}{lccc}
		\toprule
		Shape & Observation-based & Belief-based &  Receding Horizon \\
		\midrule
		Circular & $3140.5\pm130.1$  & $3154.3\pm139.9$ & $2725.6 \pm115.2$ \\ 
		T-shaped & $2601.3\pm73.2$  & $2580.7\pm66.2$ &  $2067.7 \pm 78.7$ \\
		Arc & $1789.4\pm102.2$  & $1873.9\pm112.3$  & $1646.4 \pm 103.0$ \\
		\bottomrule
	\end{tabular}
\end{table}

The inability of the observation-based approach to generalize to the arc shape stems from the reward structure differing from the true goal of the wildfire surveillance problem. Because the belief-based approach is rewarded for observing new wildfire, the learned behavior is effective in many scenarios. However, the belief-based approach does not generalize as well to different closed shapes, as demonstrated by the T-shaped wildfires. This result underscores the importance of diverse training data in order to help the network controllers generalize well to many types of fire fronts.

\section{Conclusions \label{sec:Conc}}
This paper presents a real-time guidance system for multiple fixed-wing aircraft to autonomously monitor wildfires. Given only sensor information, a deep neural network is trained to maximize wildfire surveillance for pairs of aircraft. Two approaches for handling limited sensor information were presented and evaluated through simulation. Ultimately, the trained networks could be incorporated into the on-board guidance systems of real aircraft to generate intelligent flight trajectories for monitoring wildfires. The neural network approach for image-based aircraft guidance is a general method that could be applied to other domains where information gained during flight is important for real-time trajectory generation.

\section*{Acknowledgments}
The authors wish to thank Jeremy Morton for his helpful feedback. This material is based upon work supported by the National Science Foundation Graduate Research Fellowship under Grant No. DGE-1656518. Any opinion, findings, and conclusions or recommendations expressed in this material are those of the authors and do not necessarily reflect the views of the National Science Foundation.

\bibliographystyle{IEEEtran}
\bibliography{references}
\end{document}